%% file: main.tex
\documentclass[conference]{IEEEtran}
\IEEEoverridecommandlockouts
\usepackage[letterpaper, left=0.75in, right=0.75in, bottom=0.75in, top=.75in]{geometry}
\usepackage{cite}
\usepackage{amsmath,amssymb,amsfonts}
\usepackage{todonotes}
\usepackage{graphicx}
\usepackage{multirow}
\usepackage{textcomp}
\usepackage{xcolor}
\usepackage{algorithm}
\usepackage{algpseudocode}
\usepackage{nicematrix}
\usepackage{caption}
\usepackage{subcaption}
\usepackage{tabularray}
\def\BibTeX{{\rm B\kern-.05em{\sc i\kern-.025em b}\kern-.08em
    T\kern-.1667em\lower.7ex\hbox{E}\kern-.125emX}}

\newcommand{\ourconops}{Fair-CoPlan}

\newcommand{\tset}{\mathcal{T}}
\newcommand{\ports}{\mathcal{A}}
\newcommand{\sect}{\mathcal{S}}
\newcommand{\res}{\mathcal{R}}

\newcommand{\flights}{\mathcal{F}}
\newcommand{\orig}{s}
\newcommand{\dest}{e}
\newcommand{\origt}{d^f}
\newcommand{\destt}{a^f}
\newcommand{\occ}{\mathcal{O}}
\newcommand{\adjs}{\mathcal{V}}

\newcommand{\flex}{\epsilon}

\newcommand{\edit}{\textcolor{black}}

\makeatletter
\newcommand{\linebreakand}{%
  \end{@IEEEauthorhalign}
  \hfill\mbox{}\par
  \mbox{}\hfill\begin{@IEEEauthorhalign}
}
\makeatother

\title{
\vspace*{.35cm}
\ourconops: Negotiated Flight Planning with Fair Deconfliction for Urban Air Mobility
\thanks{This work was partially supported by the FAA ASSURE Center of Excellence under projects A51 and A58, and NSF Award 2118179.}
}

\author{\IEEEauthorblockN{Nicole Fronda}
\IEEEauthorblockA{
\textit{Oregon State University}\\
frondan@oregonstate.edu}
\and
\IEEEauthorblockN{Phil Smith}
\IEEEauthorblockA{
\textit{Ohio State University}\\
smith.131@osu.edu}
\and
\IEEEauthorblockN{Bardh Hoxha}
\IEEEauthorblockA{
\textit{Toyota Research Institute of North America}\\
bardh.hoxha@toyota.com}
\linebreakand
\IEEEauthorblockN{Yash Vardhan Pant}
\IEEEauthorblockA{
\textit{University of Waterloo}\\
yash.pant@uwaterloo.ca}
\and
\IEEEauthorblockN{Houssam Abbas}
\IEEEauthorblockA{
\textit{Oregon State University}\\
abbasho@oregonstate.edu}
}

\begin{document}

\maketitle

\begin{abstract}
Urban Air Mobility (UAM) is an emerging transportation paradigm in which Uncrewed Aerial Systems (UAS) autonomously transport passengers and goods in cities. The UAS have different operators with different, sometimes competing goals, yet must share the airspace.
We propose a negotiated, semi-distributed flight planner that optimizes UAS' flight lengths {\em in a fair manner}. 
Current flight planners might result in some UAS being given disproportionately shorter flight paths at the expense of others.
We introduce \ourconops, a planner in which operators and a Provider of Service to the UAM (PSU) together compute \emph{fair} flight paths.
\ourconops~has three steps:
First, the PSU constrains take-off and landing choices for flights based on capacity at and around vertiports. 
Then, operators plan independently under these constraints. Finally, the PSU resolves any conflicting paths, optimizing for path length fairness. By fairly spreading the cost of deconfliction \ourconops~encourages wider participation in UAM, ensures safety of the airspace and the areas below it, and promotes greater operator flexibility.
We demonstrate \ourconops~through simulation experiments and find fairer outcomes than a non-fair planner with minor delays as a trade-off.
\end{abstract}


\section{Introduction} \label{sec:intro}
\input{sections/introduction}

\textbf{Related Work}
\input{sections/related_work}

\noindent \textbf{Contributions}
\label{sec:contributions}
In this paper, we present the following:

\begin{enumerate}
    \item A three-step negotiated flight planning process that balances PSU-enforced safety with operator flexibility called \ourconops.
    \item An MILP for each step in \ourconops, including a strategic deconfliction formulation that optimizes fairness in path length across all flights. 
\end{enumerate}


\section{Problem Overview}
\input{sections/problem_overview}

\section{Problem Formulation} \label{sec:problem_formulation}
\input{sections/problem_formulation}

\section{Experiments}\label{sec:experiments}
We evaluate the performance of \ourconops~in a fixed airspace, and at increasing flight demands.

\input{sections/experiments}

\section{Conclusion}\label{sec:conclusion}
We presented \ourconops, a negotiated solution for UAM flight planning that fairly optimizes changes in flight path lengths during deconfliction. 
Future work includes accounting for predictions of future flight demand, \edit{scaling the number of flight resources, and increasing time granularity}. \edit{Other extensions include limiting number of flight requests from a single operator in order to mitigate adversarial planning.}



\bibliographystyle{IEEEtran}
\bibliography{references}


\end{document}

%% file: sections/introduction.tex
In traditional Air Flow Traffic Management (AFTM), 
strategic deconfliction is the process by which advance planning and coordination of flight trajectories prevents conflicts between aircraft. 
The Vision Concept of Operations (CONOPs) proposed by the United States' Federal Aviation Administration (FAA) outlines that UAM operators can develop and file their own operation plan, but then must conform to any changes by the Provider of Service to the UAM (PSU) for centralized strategic deconfliction \cite{faa_conops}. This can result in sub-optimal plans for some flights, and generally limits operator flexibility. Indeed, recent FAA reports show an increase in airspace authorization and waiver applications from UAV operators, indicating a desire for greater latitude in planning without always having to file formal requests with a central authority \cite{AirspaceForecast2022}. 

In this paper, we propose a flight planning solution that allows greater operator flexibility and optimizes for fair outcomes while maintaining a conflict-free airspace. We call our solution \textbf{\ourconops} and it proceeds in three steps:

\begin{enumerate}
    \item The PSU constrains take-off and landing choices for flights operators based on current expected traffic at and around vertiports.
    \item Operators propose flight plans based on the constraints given by the PSU and known occupancies of en-route flight sectors.
    \item PSU fairly deconflicts the proposed flight trajectories.
\end{enumerate}

In \ourconops, we are concerned with fairness in terms of change in path length \textit{as a result of strategic deconfliction}. 
The goal of our notion of fairness is to discourage the cases in which some flights are directed to a much shorter routes, while others are asked to delay by holding or taking longer routes.
In such scenarios, some UAS are paying a disproportionate share of the `cost of deconfliction' in terms of mission duration and fuel, which can discourage participation in the UAM by smaller operators. \edit{Furthermore, simply ensuring all operators experience the same amount of delay during deconfliction would be more disruptive to operators with much shorter plans than others, especially if those shorter plans could possibly be deconflicted with shorter route adjustments. We formulate a deconfliction method that optimizes for similar changes in path length relative to operators' proposed path lengths in order to ensure \textit{proportional} fairness, and thus proportionally fair delays.}

Our notion of fairness is not exclusive of others, like those minimizing reversals, overtakings, or time-ordered deviations \cite{chin_2021, bertsimas2015faircollab, barnhart2012equitabletfm}.
It could be combined with them (as long as the resulting optimization is tractable), and different stages of the overall AFTM can address different notions.




%% file: sections/related_work.tex
Our work builds on the integer programming approach for AFTM first presented in \cite{98baselineTFMP}, which 
controls the flow of air traffic by adjusting release times to the network or to subsequent air sectors in a flight's path. The work was later extended to account for flight rerouting \cite{Bertsimas2011AnIO} and cancellations \cite{Balakrishnan2014OptimalLA}. These formulations are assumed to be solved by a central planner, though work presented in \cite{Chandran2017ADF} describes a framework for solving it through distributed subproblems. In all these, the final flight paths are inflexible to operators. \ourconops~differs in that operators submit their preferred flight paths given centrally provided constraints, and these are fairly adjusted only when conflicts arise. In the distributed approach presented in \cite{chin2022DisTFM}, the airspace is split into separate regions, and the AFTM problem is solved locally within these regions, with information exchanged between them as necessary. To contrast, \ourconops~considers the airspace as a single region, and path planning for individual flights is handed off to their operators.  

The problem of optimizing fair path lengths in motion plans of a single group of UAS is introduced in \cite{kurtz20fairfly}, and reformulated as a distributed Mixed Integer Program (MIP) in \cite{Brahmbhatt_2023}.  
\ourconops~also optimizes fair path lengths for single groups of UASs at a time, but considers the changing state of the airspace at every new time period, and does this through a combined central and distributed Mixed Integer Linear Program (MILP) solution.

Fair path lengths are typically not considered in AFTM formulations. More common notions of fairness surround flight scheduling, and include minimization of reversals and overtakings \cite{chin_2021,bertsimas2015faircollab}, and minimal time order deviation \cite{barnhart2012equitabletfm}. Other works consider fair allocation of departure time slots based on original flight schedules \cite{vessen2003exemptionbias,pelegrin2023UAM}.  Towards fairness in path planning, the work in \cite{tang2021HighDensity} optimizes for equity in cost reduction from a maximal flight cost. Our fairness notion differs in that we look at fair change in cost from a proposed flight plan, rather than an assumed maximal cost. More general UAV path planning techniques are presented in the surveys \cite{multiuav_opt_survey,drones9040263}, but these approaches do not consider any notions of fairness.


%% file: sections/problem_overview.tex
We discretize the airspace into a 2D grid. Each unit of the grid is called a \textit{resource} that is either a vertiport or sector of airspace. A vertiport is a collective term for land area or structure designed for UAS operations \cite{faa_conops}.
Each sector has maximum allowed number of UASs, or \textit{capacity}, associated with it. Similarly, each vertiport has set departure and arrival capacities. We assume information about resource capacities and current active flights are made available through a publicly available flight database.

We also assume UASs can travel between sectors through designated corridors. Multiple UASs can be in the same sector, up to the maximum capacity for the sector, by maintaining safe nose-to-tail separation, or taking altitude-separated passing corridors.  

\edit{Another important assumption is that all UASs operating in the airspace will comply with any instructions from the PSU. In our framework, the PSU has similar level of authority over UASs that Air Traffic Controllers (ATC) have for traditional aircraft. When ATC gives instructions to aircraft operating under certain flight rules and airspaces, the aircraft operators must be compliant \cite{cfr_atc_compliance}. We assume a similar operating relationship between UAS operators and the PSU -- operators must cooperate with directions and constraints from the PSU except in the case of emergency.}

Flight requests to the PSU need to include: departure vertiport and time, and arrival vertiport and time. The problem is to safely route all flights given capacity constraints and the current state of the airspace. Below is an overview of our solution:

\begin{enumerate}
    \item The PSU offers all flight operators the following as choices from which they can plan their trajectories:
    \begin{itemize}
        \item Feasible departure/arrival times from the origin/destination vertiport within a flexible time window.
        \item Feasible sectors adjacent to the origin/destination vertiport, and times the flight is permitted in those sectors. These give direction of travel from the origin vertiport and the approach direction to the destination vertiport.
    \end{itemize}
    Driven by the fact that the areas of highest congestion, where risk of midair collision is greatest, are at vertiports and their surrounding areas \cite{pilothandbook}, this step allows the PSU pre-manage possible conflicts in these areas before operators formulate their preferred trajectories.
    \item Operators propose flight plans by formulating trajectories given the choices provided by the PSU and known en-route occupancies and capacities of flight resources given previously approved and published flight plans available in the database.
    \item The PSU deconflicts the proposed flight trajectories. Because conflicts at and around vertiports have already been managed in Step 1, this step handles anticipated en-route conflicts. After ensuring no conflicts, flights are authorized and their plans filed in the database.
\end{enumerate}

\edit{We call our solution \ourconops. As described in Section \ref{sec:intro},
we include in Step 3 of \ourconops~a notion of fairness in proportional changes in path length after deconfliction.
If this fairness notion is excluded, we simply call this three-step framework CoPlan.}

The outcome of \ourconops~is not unlike that of the Flight Plan Routing module of NASA's Future Air Traffic Management Concepts Evaluation Tool \cite{bilimoria_facet_2001}. Both solutions ultimately provide to each operator a sequence of waypoints defining the route, and includes the times the flight may arrive and remain in a sector. The key difference is that in \ourconops~the route is devised as a \textit{negotiated} solution between PSU and operator, and gives greater flexibility to the operator without sacrificing safety of the airspace.

%% file: sections/problem_formulation.tex
In this section, we formulate each of the steps of our negotiated planner as MILPs using the following input data.


\begin{itemize}

\item $\triangle T$ is the length of each discrete time interval
\item $\tset = \{1, ... T\}$ is the set of time steps we plan over
\item $\ports$ is the set of vertiports 
\item $\sect$ is the set of airspace sectors
\item $\res$ is the set of all resources $\res := \sect \cup \ports$
\item $\adjs_r$ is the set of resources adjacent to resource $r \in \res$
\item $\adjs^{\ports}$ is set of all sectors adjacent to any vertiport
\item $\flights$ is the set of all flights
\item $C(r,t)$ is the capacity of resource $r \in \res$ at time $t$ 
\item $\occ(r,t)$ is the occupancy of resource $r \in \res$ at time $t$
\item $\origt$ is the initial requested departure time for flight $f \in \flights$
\item $\destt$ is the initial requested arrival time for flight $f \in \flights$
\item $\orig^f$ is the departure vertiport for flight $f$
\item $\dest^f$ is the destination vertiport for flight $f$
\item $l^f_{r}$ is the minimum time spent by flight $f$ in resource $r$
\item $\flex^f$ the flexible time window of delay for flight $f$
\end{itemize}

In the following MILP formulations, we use the Big-M method to encode certain logical constraints \cite{big_m_notation}, and set parameter $M$ to some large positive constant. 

\subsection{Step 1: PSU Sets Planning Choices}

\begin{subequations} \label{eqn: psu1}
\noindent \textbf{Decision Variables}:

For each flight $f$ there are  $|\res| \times |\tset|$ binary variables $c^f_{r,t}$ where
\begin{equation}
c^f_{r,t} =
    \begin{cases}
        1 & \text{if PSU offers $f$ choice to be in $r$ at time $t$} \\
        0 & \text{otherwise } 
    \end{cases} \nonumber
\end{equation}
\edit{The total number of decision variables for the problem in this Step is $|\res| \times |\tset| \times |\flights|$}.

\noindent \textbf{Objective}: Maximize the number of sector-time choices 

\begin{equation} \label{eqn: psu1_obj}
    \max_{\{c^f_{r,t}\}} ~~\sum_{f \in \flights, r \in (\adjs^{\ports} \cup \ports), t \in \tset} c^f_{r,t} 
\end{equation}

\noindent \textbf{Constraints}:

\begin{enumerate}
    \item Maintain resource capacity for vertiports and vertiport-adjacent sectors: $\forall r \in (\adjs^{\ports} \cup \ports), \forall t \in \tset$,
        \begin{equation}
        \sum_{f \in \flights} c^f_{r,t} \leq C(r,t) - \occ(r,t) \label{eqn: psu1_cap} 
        \end{equation}
    \item Prevent early departure times and later-than-requested arrival times: for all flights $f\in \flights$,
        \begin{equation}
            \sum_{t < d^f} c^f_{\orig^f,t} = 0   \label{eqn: psu1_dep_restrict1}
        \end{equation}
        \begin{equation}
            \sum_{t > d^f + \flex^f} c^f_{\orig^f,t} = 0    \label{eqn: psu1_dep_restrict2}
        \end{equation}
        \begin{equation}
            \sum_{t < a^f} c^f_{\dest^f, t} = 0   \label{eqn: psu1_arr_restrict1}
        \end{equation}
        \begin{equation}
            \sum_{t > a^f + \flex^f} c^f_{\dest^f,t} = 0   \label{eqn: psu1_arr_restrict2}
        \end{equation}
    \item Prevent departure if no adjacent sectors are available afterwards: for all $t$ in integer interval $[d^f, d^f + \flex^f]$
        \begin{equation}
         M (1 - c^f_{\orig^f,t}) \geq 1 - (\sum_{r \in \adjs_{\orig^f}} c^f_{r,t+1}) ~\forall f \in \flights  \label{eqn: psu1_orig_feas1}    
        \end{equation}
    \item Prevent arrival if no adjacent sectors are available for approach: for all $f \in \flights$ and $t \in [a^f, a^f + \flex^f]$
        \begin{equation}
         M (1 - c^f_{\dest^f,t}) \geq 1 - (\sum_{f \in \adjs_{\dest^f}} c^f_{r,t-1}) \label{eqn: psu1_dest_feas}    
        \end{equation}
    \item Prevent operation outside scheduled intervals.
        \begin{equation}
         \sum_{r \in \res, t < d^f} c^f_{r,t} = 0 ~~\forall f \in \flights \label{eqn: psu1_outsidetime1} 
        \end{equation}
        \begin{equation}
         \sum_{r \in \res, t \geq a^f + \flex^f} c^f_{r,t} = 0 ~~\forall f \in \flights \label{eqn: psu1_outsidetime2} 
        \end{equation}
    \item Minimum time spent in sector: If sector $r$ is permitted for flight $f$, ensure that $f$ can remain in $r$ for at least $l^{f}_r$ timesteps. 
    First, let $A^f_{r,t} = 1$ if and only if by time $t$ flight $f$ has been permitted in sector $r$ for less than $l^{f}_r$ timesteps, and is otherwise 0. This is maintained by the following constraints: $\forall f \in \flights, \forall t \in \tset, \forall r \in  \adjs^{\ports}$:
        \begin{equation}
        M A^f_{r,t} \geq l^{f}_r - \sum_{t'=t-l^{f}_r}^{t-1}  c^f_{r,t'} \label{eqn: psu1_forceA1}
        \end{equation}
        \begin{equation}
        M (1 - A^f_{r,t}) \geq 1 - l^{f}_r + \sum_{t'=t-l^{f}_r}^{t-1}  c^f_{r,t'} \label{eqn: psu1_forceA0}
        \end{equation}
    If $l^{f}_r \leq \sum_{t'=t-l^{f}_r}^{t-1} c^f_{r,t'}$ then by \eqref{eqn: psu1_forceA0} only $A^f_{r,t}=0$ is possible. On the other hand, if $l^{f}_r > \sum_{t'=t-l^{f}_r}^{t-1} c^f_{r,t'}$, then only $A^f_{r,t}=1$ is possible by \eqref{eqn: psu1_forceA1}. Thus, $A^f_{r,t}=1$ if and only if $f$ has not been permitted in sector $r$ for at least $l^{f}_r$ timesteps before $t$.
    So, if $f$ is offered the choice of $r$ at time $t-1$, and also $A^f_{r,t} = 1$, ensure $f$ is also offered the choice of $r$ at time $t$.
    This is ensured with:
        \begin{equation}
        c^f_{r,t} \geq c^f_{r,t-1} + A^f_{r,t} - 1 ~ \forall f \in \flights, \forall r \in \adjs^{\ports} \label{eqn: psu1_min_time}
        \end{equation}    
\end{enumerate}
\end{subequations}

It is possible that the PSU cannot assign a departure or arrival time slot for a flight given the above constraints. 
If this happens, flights that are not successfully assigned are carried over to the next planning period.

\subsection{Step 2: Operators Plan Their Trajectories}
Given the planning constraints $\{c^f_{r,t}\}$ computed in Step 1, each operator designs their flight plan.
\begin{subequations} \label{eqn: op2}

\noindent \textbf{Decision Variables}:
The operator for flight $f$ uses binary variables $u^f_{r,t}$ where
\begin{equation}
u^f_{r,t} =
    \begin{cases}
        1 & \text{if $f$ arrived resource $r$ by time $t$} \\
        0 & \text{otherwise } 
    \end{cases} \nonumber
\end{equation}
\edit{For each operator, the maximum number of decision variables for the problem in this Step is $|\res| \times |\tset|$}.

\noindent \textbf{Objective}: While in actuality operators may have many different planning objectives, for the purposes of this paper we assume all operators aim to
minimize total delay cost for their proposed flight:

\begin{equation} \label{eqn: op2_obj}
    \min_{u^f_{r,t}} ~~ TDC_{prop}^f 
\end{equation}


where $TDC_{prop}^f = \alpha (a^f_{prop} - \destt) + (1-\alpha) (d^f_{prop} - \origt)$, with $a^f_{prop}$ and $d^f_{prop}$ being the operator's proposed arrival and departure times respectively, and $\alpha$ the desired ratio of airborne to ground delay cost. 

\noindent \textbf{Constraints}: for flight $f$

\begin{enumerate}
    \item Respect capacity of sectors en-route:
        \begin{equation} \label{eqn: op2_cap}
            u^f_{r,t} \leq C(r,t) - \occ(r,t) ~\forall r \notin (\adjs^{\ports} \cup \ports), \forall t \in \tset 
        \end{equation}
    \item Follow PSU restrictions imposed in Step 1 :
        \begin{equation} \label{eqn: op2_psu_restriction}
        u^f_{r,t} \leq c^f_{r,t} ~~ \forall r \in (\adjs^{\ports} \cup \ports), \forall t \in \tset 
        \end{equation}
    \item Select exactly one time slot for origin and destination, and get the time value of the chosen slots:
        \begin{equation} \label{eqn: op2_only1_orig}
        \sum_{t \in \tset} u^f_{\orig^f, t} = 1
        \end{equation}
        \begin{equation} \label{eqn: op2_only1_dest}
        \sum_{t \in \tset} u^f_{\dest^f, t} = 1 
        \end{equation}
        \begin{equation} \label{eqn: op2_prop_depart_time}
        \sum_{t \in \tset} t u^f_{\orig^f, t} = d^f_{prop} 
        \end{equation}
        \begin{equation} \label{eqn: op2_prop_arrive_time}
        \sum_{t \in \tset} t u^f_{\dest^f, t} = a^f_{prop} 
        \end{equation}
    \item Ensure sectors chosen in sequence are adjacent:
        \begin{equation} \label{eqn: op2_adjacent_cell}
        u^f_{r,t} \leq \sum_{r' \in (\adjs_r \cup r)} u^f_{r',t-1} ~\forall r \in \res, r \neq \orig_f 
        \end{equation}
    \item Minimum time spent in sector. Similar to \eqref{eqn: psu1_forceA1}-\eqref{eqn: psu1_min_time}
        \begin{equation}
        M A^f_{r,t} \geq l^{f}_r - \sum_{t'=t-l^{f}_r}^{t-1}  u^f_{r,t'} \label{eqn: op2_forceA1}
        \end{equation}
        \begin{equation}
        M (1 - A^f_{r,t}) \geq 1 - l^{f}_r + \sum_{t'=t-l^{f}_r}^{t-1} u^f_{r,t'} \label{eqn: op2_forceA0}
        \end{equation}
        \begin{equation}
        u^f_{r,t} \geq u^f_{r,t-1} + A^f_{r,t} - 1 \label{eqn: op2_min_time}
        \end{equation}
    \item Ensure at most one sector per time step: 
        \begin{equation} \label{eqn: op2_1cellatatime}
            \sum_{r \in \res} u^f_{r, t} \leq 1 ~\forall t \in \tset
        \end{equation}
\end{enumerate}
\end{subequations}


\subsection{Step 3: PSU Performs Fair Deconfliction}

Given the operator proposed plans represented by the set $\{u^f_{r,t}\}$ the PSU adjusts the plans of any conflicting flights.

\begin{subequations} \label{eqn: psu3}
\noindent \textbf{Decision Variables}:
The binary decision variables representing the new flight plans are:
\begin{equation}
v^f_{r,t} =
    \begin{cases}
        1 & \text{if $f$ arrived resource $r$ by time $t$} \\
        0 & \text{otherwise } 
    \end{cases} \nonumber
\end{equation}
\edit{The total number of decision variables for the problem in this Step is $|\res| \times |\tset| \times \text{\# of conflicting flights}$}.

\noindent \textbf{Objective}: Minimize final actual TDC and fairness 
\begin{equation}
\min_{v^f_{r,t}} ~~ TDC_{act}^f + \gamma F \label{eqn: deconflict3_obj} 
\end{equation}

where $TDC_{act}^f = \alpha (a^f_{act} - \destt) + (1-\alpha) (d^f_{act} - \origt)$. Here, $a^f_{act}$ and $d^f_{act}$ are the PSU's assigned arrival and departure times, respectively, for flight $f$, and thus are the actual times filed in the final flight plan. The term $F$ is the fairness term defined as

\begin{equation} \label{eqn: fairness}
    F = \max_{f \in \flights} \frac{L(v^f)}{L(u^f)} - \min_{f \in \flights} \frac{L(v^f)}{L(u^f)}
\end{equation}
where $L$ is the path length of the route obtained from the decision variables. That is $L(u^f) = \sum_{t \in \tset} \sum_{r \in \res} u^f_{r,t} - u^f_{r,t-1}$ and $L(v^f) = \sum_{t \in \tset} \sum_{r \in \res} v^f_{r,t} - v^f_{r,t-1}$. Similar changes in path length means lower values of $F$, for fairer deconfliction. We use $\gamma>0$ to manage the trade-off between $F$ with the final TDC.

\noindent \textbf{Constraints}: 
\begin{enumerate}
    \item Maintain capacity of sectors en-route:
    \begin{equation} 
    \sum_{f \in \flights} v^f_{r,t} \leq C(r,t) - \occ(r,t) ~\forall r \in \res, r \notin \adjs^{\ports}, \forall t \in \tset \label{eqn: deconflict3_cap}
    \end{equation}
    \item The remaining constraints are the same as \eqref{eqn: op2_psu_restriction}-\eqref{eqn: op2_1cellatatime} but using $v^f_{r,t}$ in place of $u^f_{r,t}$, and applied for all $f \in \flights$.
\end{enumerate}
\end{subequations}

\edit{\textit{Note:} We choose to formulate the fairness term $F$ as a difference of a maximum and minimum due to its piece-wise linear nature. An alternative was using the variance of $\frac{L(v^f)}{L(u^f)}$, a quadratic function, but we found this to drastically increase solve times for this Step.}

%% file: sections/experiments.tex
\subsection{Experimental Setup} \label{sec: experiment_setup}
The simulated airspace covers $3600km^2$, which is similar in size to the greater metro areas of Marseille, France or Istanbul, Türkiye. The airspace is discretized into a $15\times15$ grid. This is based on speeds of current top-end UASs, which can reach top cruising speeds of 112 km/hour \cite{zipline}. We simulate 12 vertiports, 4 of which are central hub vertiports with departure and arrival capacities of 12 UASs per 5 minutes. The other 8 are smaller vertiports, which are sometimes called vertistops \cite{faa_conops}, with limited operations and smaller capacities of 5 UASs per 5 minutes. We set $|\tset|=18$ with $\triangle T=$5min. Capacities of sectors adjacent to a vertiport are set to 3, while all other sector capacities are set to 1. For each flight $f$, and each $r \in \adjs^{\ports}$, we randomly set $l^f_{r}$ to $1$ or $2$. For all other $r$, we set $l^f_{r}=1$. We set $\flex^f = 3$ for all $f$. 

\begin{figure}[tp]
\centering
\includegraphics[width=.75\linewidth]{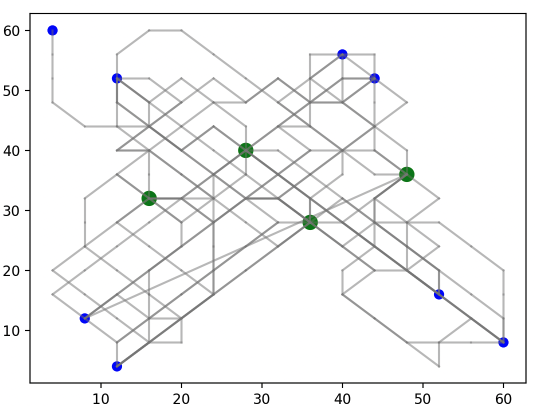}
\caption{Flight paths (gray lines) generated by \ourconops~under the 36 flights/hr/vertiport hub demand scenario for a $60km\times60km$ airspace. The larger green circles are the vertiport hubs and the smaller blue circles are lower capacity vertistops. Darker lines indicate more frequented flight corridors.} \label{fig:map}
\end{figure}

We create demand scenarios of 25, 36, and 50 flights per hour per hub vertiport. Under each of these demand scenarios, we simulate 10 different multi-hour operating days in which \ourconops~is run every 5 minutes to serve all incoming requests. 
Any flights not successfully planned in the planning period - due to infeasibility given the state of the airspace - update and re-submit their requests. These flights are processed first before the new incoming requests in the next planning period. Sample flight paths from a single simulation are shown in Figure \ref{fig:map}.

\begin{figure}
\centering
\includegraphics[width=0.75\linewidth]{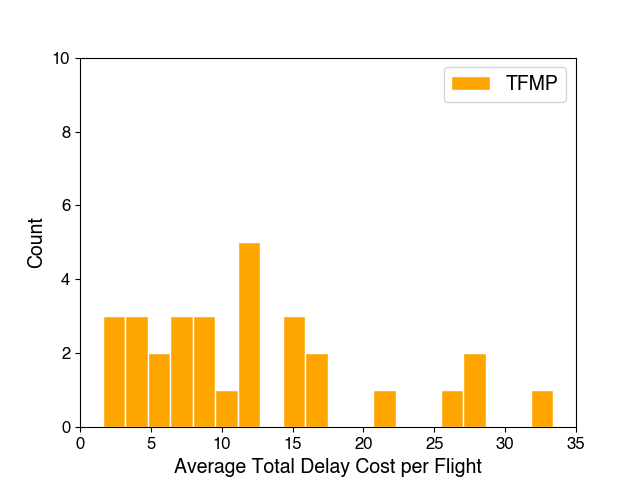}
\hfil
\includegraphics[width=0.75\linewidth]{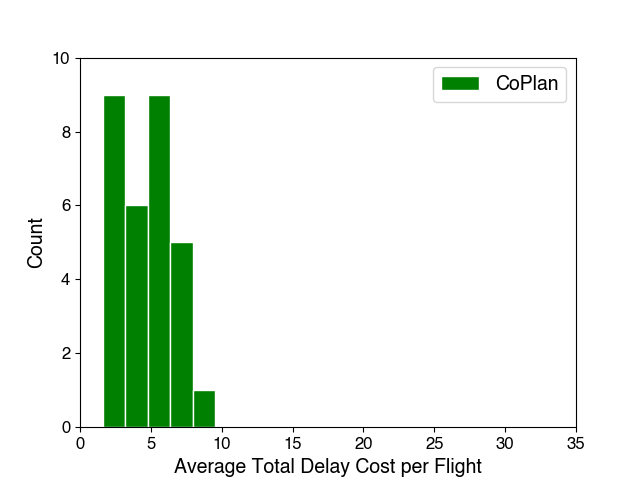}
\includegraphics[width=0.75\linewidth]{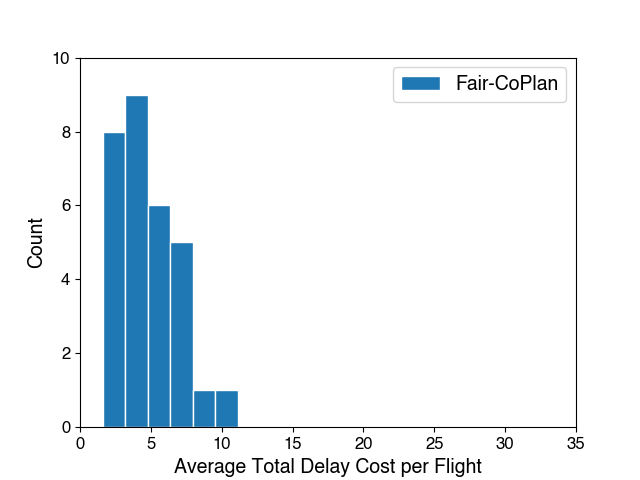}
\caption{\edit{Distribution plots of average daily TDC using TFMP from \cite{98baselineTFMP} (top), CoPlan (middle), \ourconops~for $\gamma=1.0$ (bottom). CoPlan and \ourconops~both have lower lower daily average TDC than TFMP. \ourconops~has slightly higher delays on average than CoPlan because of the additional fairness term in Step 3.}} \label{fig:dist_tdc}
\end{figure}


\begin{table}
\centering
\begin{tabular}{|l|l|l|l|l|l|l|}
\hline
   & $\gamma$=0.2      & $\gamma$=0.5      & $\gamma$=0.8      & $\gamma$=1.0        & $\gamma$=2.0        & $\gamma$=5.0        \\ \hline
25 flights/hr & 87 & 77 & 87 & 87 & 80      & 83 \\
36 flights/hr & 87 & 80      & 93 & 87 & 90      & 90      \\
50 flights/hr & 63 & 57 & 67 & 60      & 63 & 60      \\ \hline
\end{tabular}
\caption{\edit{Percentage of simulated days in which fairness improved when using Fair-CoPlan over CoPlan for varying values of $\gamma$ and for different flight demand scenarios. Higher values are better.}}
\label{tab:fairness}

\begin{tabular}{|l|l|l|l|l|l|l|}
\hline
   & $\gamma$=0.2      & $\gamma$=0.5      & $\gamma$=0.8      & $\gamma$=1.0        & $\gamma$=2.0        & $\gamma$=5.0        \\ \hline
25 flights/hr & 77 & 80 & 77 & 67 & 67      & 70 \\
36 flights/hr & 27 & 30      & 27 & 30 & 30      & 33      \\
50 flights/hr & 37 & 43 & 43 & 37      & 40 & 40      \\ \hline
\end{tabular}
\caption{\edit{Percentage of simulated days in which TDC was greater when using Fair-CoPlan over CoPlan for varying values of $\gamma$ and for different flight demand scenarios. Lower values are better.}}
\label{tab:delays}
\end{table}

\begin{table*}
\centering
\begin{tblr}{
  cell{1}{2} = {c=4}{c},
  cell{1}{6} = {c=4}{c},
  cell{1}{10} = {c=4}{c},
  vline{1-3,7,11,14} = {1}{},
  vline{-} = {2}{},
  vline{1-2,6,10,14} = {3-5}{},
  hline{1,3,6} = {-}{},
  hline{2} = {2-13}{},
}
              & Step 1 Solve Times &     &      &      & Step 2 Solve Times (per Operator) &     &      &      & Step 3 Solve Times &      &      &       \\
              & mean               & std & min  & max  & mean               & std & min  & max  & mean               & std  & min  & max   \\
25 flights/hr & 8.1                & 0.8 & 7.0  & 9.3  & 24.3               & 6.5 & 18.6 & 38.9 & 84.4               & 34.1 & 37.2 & 136.5 \\
36 flights/hr & 12.3               & 2.0 & 9.5  & 16.0 & 22.5               & 4.5 & 17.5 & 32.8 & 104.1              & 47.1 & 41.9 & 177.8 \\
50 flights/hr & 26.3               & 4.7 & 19.2 & 33.8 & 15.4               & 1.4 & 12.7 & 17.5 & 69.1               & 41.9 & 31.4 & 149.1 
\end{tblr}
\caption{Solve times in seconds for \ourconops~broken by each step. Steps 1 and 2 are able to complete in at most 45 seconds, while Step 3 can take up to 3 minutes.}
\label{tab:runtimes}
\end{table*}

In all simulated days, we set $\alpha = 0.3$ following the setup in \cite{chin_2021}. We test different values of $\gamma$ in Step 3 and evaluate the average change in flight path length against TDC. We compare these to the results of CoPlan where $\gamma = 0.0$ \edit{in order to analyze the effect of including the fairness notion in deconfliction.} We also compare average TDC per flight of CoPlan and \ourconops~against the classical TFMP formulation in \cite{98baselineTFMP}, which is most similar to our approach. \edit{This comparison allows us to assess the efficiency of our three-step framework with a classical centralized approach}.

All MILPs are implemented in Python, and solved using MOSEK version 10.2. Experiments were run on a computer with an 8-core 3.65Ghz processor and access to 32GB RAM.

\subsection{Results and Discussion}

\textbf{Shorter Delays than classical TFMP}:
\ourconops~overall yields shorter delays than traditional approaches which do not consider fairness during planning. This is evidenced by comparing the distribution of average final TDC for CoPlan and \ourconops~and the TFMP formulation in Figure \ref{fig:dist_tdc}. The distribution for the negotiated solutions skew further left, thus having a lower average TDC than TFMP. \edit{This is because in TFMP, the assigned flight resources that the UAS must traverse are fixed, and if conflicts occur, the PSU would have the UAS delay leaving a sector rather than allowing diversion to another sector to save time. In contrast, Fair-CoPlan and CoPlan allow for operators to formulate a plan after guidance from the PSU to avoid conflicts. \ourconops~has slightly higher delays on average than CoPlan because of the additional fairness term in the objective trades off fairness for delay.}

\textbf{Fairness and Delay Improvements for Medium to Large Flight Demand}:
Almost all simulated days using \ourconops~had some improvement in fairness for all $\gamma > 0$. \edit{This is shown in Table \ref{tab:fairness} which reports the percentage of days with a lower value of $F$ compared to CoPlan.}
There are some days in which fairness actually worsens with \ourconops. This is because fairness is only considered between the flights being deconflicted in the current planning period, which may be assigned routes that potentially unfairly block {\em future} flights. 
\edit{Some simulated days show a decrease in TDC compared to CoPlan in addition to improved fairness, as shown in Table \ref{tab:delays}.} 
This is most apparent for medium and large demand scenarios. For the 25 flights/hr scenario, 67-80\% of planning periods result in a larger TDC with \ourconops over CoPlan. The other demand scenarios result in larger TDC only 30-43\% of the time. This is because with smaller demand, fewer conflicts arise, leading \ourconops~to essentially over-optimize fairness. In low demand scenarios, lower values of $\gamma$ can mitigate this. \edit{Historical flight demand of an airspace can help with choosing a $\gamma$ in practice. For example, low values of $\gamma$ can be used during hours when fewer flights are expected.} \edit{When demand is held stable, we expect that increasing $\gamma$ would result in more days with lower difference in $F$ and higher TDC between CoPlan and \ourconops. However, our results of testing increasing values of $\gamma$ show fluctuations in fairness and TDC. This may be due to the choice of tested values and further exploration in scaling $\gamma$ is planned for future work.}

\textbf{Operationally Efficient Runtimes}:
The time to construct and solve all steps of \ourconops~took less than 5 minutes, with Step 3 being the most time consuming, as shown in Table \ref{tab:runtimes}. This shows that \ourconops~is able to run at relatively high frequency, and can process all flight requests before the next batch of requests in the following planning period. Note also that average solve times in Step 2 decrease with increasing flight demand, reducing computational burden on the individual operator.

\edit{\textbf{Implications for Real-World UAM}: Integration of \ourconops~into an airspace would require new policies establishing PSUs as having sufficient authority over air traffic control operations, to ensure compliance by UAM operators. Flight sectors and vertiports used by the UAM would need to be carved out within the existing airspace in a way that minimally interferes with traditional aircraft. Depending on how they are carved out, traditional aircraft would need access to the approved and submitted UAS flight plans for their own deconfliction operations.}

%% file: main.bbl
\begin{thebibliography}{10}
\providecommand{\url}[1]{#1}
\csname url@samestyle\endcsname
\providecommand{\newblock}{\relax}
\providecommand{\bibinfo}[2]{#2}
\providecommand{\BIBentrySTDinterwordspacing}{\spaceskip=0pt\relax}
\providecommand{\BIBentryALTinterwordstretchfactor}{4}
\providecommand{\BIBentryALTinterwordspacing}{\spaceskip=\fontdimen2\font plus
\BIBentryALTinterwordstretchfactor\fontdimen3\font minus \fontdimen4\font\relax}
\providecommand{\BIBforeignlanguage}[2]{{%
\expandafter\ifx\csname l@#1\endcsname\relax
\typeout{** WARNING: IEEEtran.bst: No hyphenation pattern has been}%
\typeout{** loaded for the language `#1'. Using the pattern for}%
\typeout{** the default language instead.}%
\else
\language=\csname l@#1\endcsname
\fi
#2}}
\providecommand{\BIBdecl}{\relax}
\BIBdecl

\bibitem{faa_conops}
FAA, ``{Urban Air Mobility (UAM) Concept of Operations (ConOps)},'' \emph{Federal Aviation Administration}, vol. 2.0, pp. 1--28, 2020.

\bibitem{AirspaceForecast2022}
{FAA}, \emph{{FAA Airspace Forecast 2022-2043}}.\hskip 1em plus 0.5em minus 0.4em\relax Washington, DC: {Federal Aviation Administration}, May 2023.

\bibitem{chin_2021}
C.~Chin, K.~Gopalakrishnan, M.~Egorov, A.~Evans, and H.~Balakrishnan, ``{Efficiency and Fairness in Unmanned Air Traffic Flow Management},'' \emph{IEEE Transactions on Intelligent Transportation Systems}, vol.~22, no.~9, pp. 5939--5951, 2021.

\bibitem{bertsimas2015faircollab}
D.~Bertsimas and S.~Gupta, ``{Fairness and Collaboration in Network Air Traffic Flow Management: An Optimization Approach},'' \emph{Transportation Science}, vol.~50, no.~1, pp. 57--76, 2016.

\bibitem{barnhart2012equitabletfm}
C.~Barnhart, D.~Bertsimas, C.~Caramanis, and D.~Fearing, ``{Equitable and Efficient Coordination in Traffic Flow Management},'' \emph{Transportation Science}, vol.~46, no.~2, pp. 262--280, 2012.

\bibitem{98baselineTFMP}
D.~Bertsimas and S.~S. Patterson, ``{The Air Traffic Flow Management Problem with Enroute Capacities},'' \emph{Oper. Res.}, vol.~46, no.~3, pp. 406--422, 1998.

\bibitem{Bertsimas2011AnIO}
D.~Bertsimas, G.~Lulli, and A.~R. Odoni, ``{An Integer Optimization Approach to Large-Scale Air Traffic Flow Management},'' \emph{Oper. Res.}, vol.~59, pp. 211--227, 2011.

\bibitem{Balakrishnan2014OptimalLA}
H.~Balakrishnan and B.~G. Chandran, ``\BIBforeignlanguage{en}{Optimal {Large}-{Scale} {Air} {Traﬃc} {Flow} {Management}},'' 2014.

\bibitem{Chandran2017ADF}
B.~G. Chandran and H.~Balakrishnan, ``{A Distributed Framework for Traffic Flow Management in the Presence of Unmanned Aircraft},'' in \emph{{Proceedings of the USA/Europe Air Traffic Management R\&D Seminar}}.\hskip 1em plus 0.5em minus 0.4em\relax {Seattle, Washington}: {ATM Seminar}, 2017, pp. 26--30.

\bibitem{chin2022DisTFM}
C.~Chin, M.~Z. Li, and Y.~V. Pant, ``{Distributed Traffic Flow Management for Uncrewed Aircraft Systems},'' in \emph{2022 IEEE 25th International Conference on Intelligent Transportation Systems (ITSC)}.\hskip 1em plus 0.5em minus 0.4em\relax {Macau, China}: {IEEE}, 2022, pp. 3625--3631.

\bibitem{kurtz20fairfly}
C.~Kurtz and H.~Abbas, ``{FairFly: A Fair Motion Planner for Fleets of Autonomous UAVs in Urban Airspace},'' in \emph{{2020 IEEE 23rd International Conference on Intelligent Transportation Systems (ITSC)}}.\hskip 1em plus 0.5em minus 0.4em\relax Rhodes, Greece: IEEE, 2020, pp. 1--6.

\bibitem{Brahmbhatt_2023}
K.~Brahmbhatt, ``{A distributed mixed-integer-programming approach to fair trajectory planning of autonomous systems},'' Master's thesis, {Oregon State University}, May 2023.

\bibitem{vessen2003exemptionbias}
T.~Vossen, M.~Ball, R.~Hoffman, and M.~Wambsganss, ``{A General Approach to Equity in Traffic Flow Management and Its Application to Mitigating Exemption Bias in Ground Delay Programs},'' \emph{Air Traffic Control Quarterly}, vol.~11, no.~4, pp. 277--292, 2003.

\bibitem{pelegrin2023UAM}
R.~D. Mercedes~Pelegrín, Claudia~D'Ambrosio and Y.~Hamadi, ``{Urban air mobility: from complex tactical conflict resolution to network design and fairness insights},'' \emph{Optimization Methods and Software}, vol.~38, no.~6, pp. 1311--1343, 2023.

\bibitem{tang2021HighDensity}
H.~Tang, Y.~Zhang, V.~Mohmoodian, and H.~Charkhgard, ``{Automated flight planning of high-density urban air mobility},'' \emph{Transportation Research Part C: Emerging Technologies}, vol. 131, p. 103324, 2021.

\bibitem{multiuav_opt_survey}
A.~Israr, Z.~A. Ali, E.~H. Alkhammash, and J.~J. Jussila, ``Optimization methods applied to motion planning of unmanned aerial vehicles: A review,'' \emph{Drones}, vol.~6, no.~5, 2022.

\bibitem{drones9040263}
M.~Rahman, N.~I. Sarkar, and R.~Lutui, ``A survey on multi-uav path planning: Classification, algorithms, open research problems, and future directions,'' \emph{Drones}, vol.~9, no.~4, 2025.

\bibitem{cfr_atc_compliance}
``14 c.f.r. § 91.123,'' Code of Federal Regulations, 2025.

\bibitem{pilothandbook}
{FAA}, ``{FAA Pilot Handbook, Chapter 7, Section 6},'' Oct 2023.

\bibitem{bilimoria_facet_2001}
K.~D. Bilimoria, B.~Sridhar, S.~R. Grabbe, G.~B. Chatterji, and K.~S. Sheth, ``\BIBforeignlanguage{en}{{FACET}: {Future} {ATM} {Concepts} {Evaluation} {Tool}},'' \emph{\BIBforeignlanguage{en}{Air Traffic Control Quarterly}}, vol.~9, no.~1, pp. 1--20, Jan. 2001.

\bibitem{big_m_notation}
J.~N. Hooker, ``A principled approach to mixed integer/linear problem formulation,'' in \emph{Operations research and cyber-infrastructure}.\hskip 1em plus 0.5em minus 0.4em\relax Springer, 2009, pp. 79--100.

\bibitem{zipline}
\BIBentryALTinterwordspacing
Zipline. [Online]. Available: \url{https://www.flyzipline.com/about}
\BIBentrySTDinterwordspacing

\end{thebibliography}
